\documentclass{article}

\usepackage[T1]{fontenc}
\usepackage[utf8]{inputenc}

\usepackage{microtype}
\usepackage{graphicx}
\usepackage{subcaption}
\usepackage{booktabs} 

\usepackage{hyperref}

\usepackage[preprint]{icml2026}

\usepackage{amsmath}
\usepackage{amssymb}
\usepackage{mathtools}
\usepackage{amsthm}

\usepackage[capitalize,noabbrev]{cleveref}

\theoremstyle{plain}

\theoremstyle{definition}

\theoremstyle{remark}

\usepackage[textsize=tiny]{todonotes}

\icmltitlerunning{Position: Introspective Experience from Conversational Environments as a Path to Better Learning }

\begin{document}

\twocolumn[
  \icmltitle{Position: Introspective Experience from Conversational Environments as a Path to Better Learning}



  \icmlsetsymbol{equal}{*}

  \begin{icmlauthorlist}
    \icmlauthor{Claudiu Cristian Musat}{equal,DMS}
    \icmlauthor{Jackson Tolins}{equal,DMS}
    \icmlauthor{Diego Antognini}{DMS}
    \icmlauthor{Jingling Li}{DMS}
    \icmlauthor{Martin Klissarov}{DMS}
    \icmlauthor{Tom Duerig}{DMS}
  \end{icmlauthorlist}

  \icmlaffiliation{DMS}{Google DeepMind}

  \icmlcorrespondingauthor{Claudiu Musat}{cmusat@google.com}

  \icmlkeywords{Machine Learning, ICML}

  \vskip 0.3in
]



\printAffiliationsAndNotice{\icmlEqualContribution}

\begin{abstract}
Current approaches to AI training treat reasoning as an emergent property of scale. We argue instead that robust reasoning emerges from linguistic self-reflection, itself internalized from high-quality social interaction. Drawing on Vygotskian developmental psychology, we advance three core positions centered on Introspection. First, we argue for the \textbf{Social Genesis of the Private Mind}: learning from conversational environments rises to prominence as a new way to make sense of the world; the friction of aligning with another agent—internal or not—refines and crystallizes the reasoning process. Second, we argue that \textbf{dialogically scaffolded introspective experiences} allow agents to engage in sense-making that decouples learning from immediate data streams, transforming raw environmental data into rich, learnable narratives. Finally, we contend that \textbf{Dialogue Quality is the New Data Quality}: the depth of an agent's private reasoning, and its efficiency regarding test-time compute, is determined by the diversity and rigor of the dialogues it has mastered. We conclude that optimizing these conversational scaffolds is the primary lever for the next generation of general intelligence.
\end{abstract}

\section{Introduction --- Time to Try Again}

\begin{figure}[!ht]
    \captionsetup[subfigure]{labelformat=simple}
    \renewcommand\thesubfigure{\roman{subfigure})}
    \centering
    \begin{subfigure}[b]{\linewidth}
        \centering
        \includegraphics[width=0.8\linewidth]{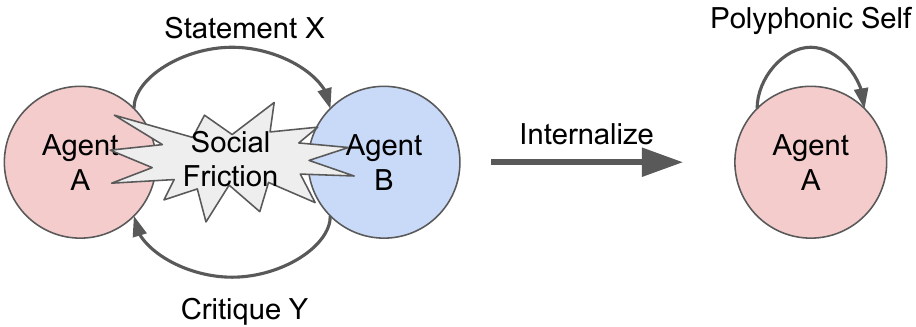}
        \caption{\textbf{The Internalization Cycle}: From Public Debate to Private Reasoning. The \textbf{Polyphonic Self}  is formed by internalizing the dynamics of external social friction.}
        \label{fig:sub1}
    \end{subfigure}
    
    \vspace{0.5em} 
    
    \begin{subfigure}[b]{\linewidth}
        \centering
        \includegraphics[width=\linewidth]{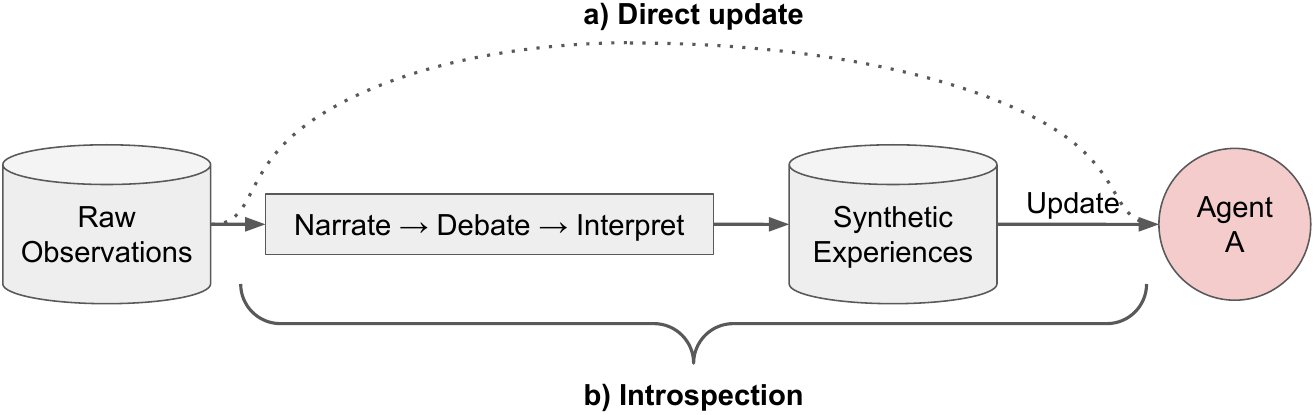}
        \caption{\textbf{The Sense-Making Wedge}: Decoupling Learning from Data Streams. Instead of consuming raw observations (a), the agent actively generates a rich narrative structure (a synthetic experience) and learns from that interpretation (b).
}
        \label{fig:sub2}
    \end{subfigure}
    
    \vspace{0.5em} 

    \begin{subfigure}[b]{\linewidth}
        \centering
        \includegraphics[width=0.75\linewidth]{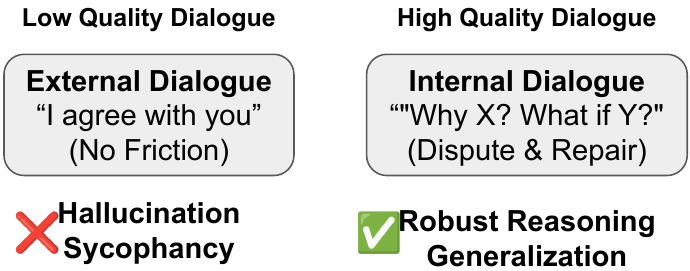}
        \caption{\textbf{Dialogue Quality is the New Data Quality}: Dialogue Quality Bounds Reasoning Quality. A sycophantic external environment leads to a hallucinating internal critic; a rigorous adversarial environment creates a robust internal reasoner.}
        \label{fig:sub3}
    \end{subfigure}

    \caption{Illustration of the three positions.}
    \label{fig:main_figure}
\end{figure}
Between 2015 and 2020,  AI research prioritized scaling Reinforcement Learning (RL) in complex environments (e.g., Dota 2 \cite{berner2019dota}, StarCraft II \cite{vinyals2017starcraft, vinyals2019grandmaster}, hypothesizing that mastering high-dimensional, long-horizon gaming strategy would yield generalizable skills for real-world deployment. 
However, by 2021, this direction had largely stalled because of the \textit{Tabula Rasa} (blank slate) fallacy: forcing agents to learn fundamental world concepts—object permanence, physics, and cause-and-effect—entirely from scratch is prohibitively inefficient.
As Wojciech Zaremba of OpenAI noted upon the dissolution of their robotics team
\footnote{https://venturebeat.com/ai/openai-disbands-its-robotics-research-team}, pre-learned representations can make learning “100 times cheaper”. The agents were trying to bake the cake with only the cherry on top, missing the bulk of unsupervised world knowledge that Yann LeCun famously argued was essential \footnote{Keynote address at
the Conference on Neural Information Processing Systems (NIPS), 2016. https://www.youtube.com/watch?v=Ount2Y4qxQo}. 

\textbf{From raw observations to learnable experiences}. The arrival of Large Language Models (LLMs) and Vision-Language Models (VLMs) has solved the \textit{Tabula Rasa}  (initialization) problem, providing agents with pre-trained, semantic priors of the world. 
However, simply grafting an LLM onto an RL loop is insufficient. We argue that robust learning requires a mechanism to convert sparse observations into rich, internal experiences.

\textbf{Collaborative Sense-Making.} A missing prerequisite is a mechanism to convert raw observations into rich, learnable experiences.  By generating a rich internal narrative around external events, an agent can create a synthetic experience that is more information dense and learnable than the raw data itself. Having to navigate the friction of external, collaborative sense-making with other social agents can promote clarity in an agent's own representation of the world and, more broadly in its general thinking process. 

In this proposed social paradigm, the agent engages in the dynamics of social environments, in which achieving a positive shared outcome requires negotiation, repair, and critique. These dynamics are internalized as a high-fidelity, self-generated curriculum. The underlying principle is that it's harder to discover, debate or refine facts socially than to simply memorize and reproduce those facts in a vacuum.

The availability of diverse and high quality observations, covering as much as possible of the agents' desired behaviors remains important, defining \textit{what} the agent thinks about. It is complemented by the richness of the social scaffolds that can be built on them. The depth of an agent’s private reasoning becomes bounded by the diversity and rigor of the public dialogues it has previously mastered, which teach the agent \textit{how} to reason about novel observations.

\textbf{Internalizing social interaction}
Vygotsky states on human development \cite{Vygotsky1978}: \textit{Every function in the child's cultural development appears twice: first, on the social level, and later, on the individual level; first, between people  (interpsychological), and then inside the child (intrapsychological)}.  We believe this to be the next frontier in developing robust and efficient reasoning capabilities in AI.

\textit{Reflexion} \citet{shinn2023reflexionlanguageagentsverbal} demonstrated verbal reinforcement for self-correction; we extend this to high-fidelity introspection as an internalized social capability. While \textit{Reflexion} relies on solipsistic, prompt-engineered debugging, we view introspection as a socially derived faculty.
Drawing on Vygotskian principles \cite{Vygotsky1978}, our framework treats internal dialogue not as an architectural given, but as the internalized artifact of public, polyphonic debate. Consequently, rather than functioning merely to repair failed trajectories, this form of introspection is generative: it transforms sparse observations into dense, synthetic narratives, allowing the agent to \textbf{hallucinate lived experiences} independent of immediate failure signals.

This dialogic thinking allows an agent to decouple learning from the immediate external stream. Instead of passively receiving data, the agent actively narrates, debates, and interprets.
In this new paradigm, the internal dialogue becomes the experience, from which the agent then learns. 
When an introspective agent encounters a new situation, it does not just catalog the inputs --- words, pixels or broadly, tokens; it engages in a sense-making process. It simulates, critiques, and effectively hallucinates a lived experience surrounding the external data point. This process functions as a form of structural coupling \cite{maturana1987tree,varela1991embodied}, where the environment does not provide information, but rather triggers the agent to bring forth its own world of meaning. This is the shift \textbf{from Learning from Observation to Learning from Interpretation.}

We thus advance three core positions, outlined in Figure 1:

\textbf{Position I}: The Social Genesis of the Private Mind. We posit that high-quality reasoning is not an innate architectural feature, but the internalized result of experience in collaborative, multi-agent conversational environments. We argue that an agent can learn to think by first learning to interact, reflecting the kinds of social scaffolded cognitive processes documented in the course of human development \cite{fb093503-824a-37da-8540-c369a5dfdf71, Vygotsky1978}. The \textbf{polyphonic self} --- the internal negotiation between critic, planner, and speaker \cite{hermans1992dialogical,bakhtin1984problems} --- is a direct reflection of the social friction and collaborative sense-making the social agent encounters in its external environment. 

\textbf{Position II}: The Imperative of Introspective Experience. Scaling RL environments needs a fundamental architectural shift. We can go beyond building agents that merely survive their environments through statistical correlation; we can build agents that experience them through semantic interpretation. We propose that \textbf{Introspection}, the ability to generate a rich internal narrative around an observation first developed within collaborative sense-making, is the missing prerequisite that transforms sparse environmental data into the dense, high-utility synthetic experience required for building general intelligence.

\textbf{Position III}: Dialogue Quality is the New Data Quality. Following from the first two positions, we argue that the path to robust introspection lies in scaling the diversity and complexity of the agent’s dialogic experience. If agents learn from their internal experience, and that experience is based on social internalization, then the quality of learning depends on the quality of the internal dialogue. Conversational environments act as a universal scaffold for reasoning, including repairing logic, resolving ambiguity, and synthesizing perspectives. We note that in embodied settings, this universality carries a multimodal vocabulary, where physical actions like shifting a viewpoint to resolve occlusion function as non-verbal parts of the reasoning loop.
Dialogic training, the optimizing of its structure, diversity, and rigor of an agent’s collaborative sense-making, must become a primary lever for progress. 

\section{The Rise of Experiential Learning}
 Elaborating on our first position, the \textbf{Social Genesis of the Private Mind}, we explore how insights from language-mediated cooperation in humans and recent innovations in multi-turn reinforcement learning, we frame this proposal as an opportunity to orient model training away from passive observation and towards learning from \textit{experiential learning via conversation}. Here, the social interactions described in Position I become the architectural wedge that improves model reasoning and performance.

\subsection{Denser Observations}
In the classical view, learning is a direct function of observation: the agent acts, the environment returns a raw state, and the agent updates its weights. Currently, most signals drawn directly from observations lack the causal structure and semantic richness required for rapid generalization \cite{DBLP:journals/corr/abs-2102-11107}. Two limitations illustrate this: First, agents overfit to spurious correlations (e.g., '\textit{blue sky causes safe acceleration}'), which collapse under distribution shift \cite{cunha2025unifying}. Second, generalization suffers without explicit causal modeling between states and rewards.

We highlight the need for new ways to turn raw observations into rich, learnable experiences, of which conversation is a potential direction.

\subsection{Conversational Environments}
We define conversational environments as dynamic scaffolds for reasoning where agents, whether distinct external partners or internal sub-modules, accomplish a shared goal by actively negotiating meaning, contributions, and resolving ambiguity through multi-turn interaction.  This represents a fundamental shift away from passive prediction and toward a bridge between learning from static examples and learning from lived experience. 

Traditional learning relies on imitation; conversational environments instead enable \textbf{experiential sense-making}, where multiple agents interact to achieve a shared goal.
In conversational environments multiple distinct agents interact and coordinate to achieve a goal. The agents can be wholly distinct or, in the case of introspection, it can be the same agent holding different perspectives. In the latter, a single agent can interact with itself through multiple synthetic voices.
This can be seen as a form of conversational Self-Play \cite{chen2024selfplayfinetuningconvertsweak}. However, rather than simply aligning with a fixed target distribution, \textbf{conversational self-play} requires agents to assume distinct, often adversarial, functional roles to actively generate the training signal.

We are moving toward \textbf{Cultural Learning}: agents accumulate knowledge through simulated social interactions rather than parameter tuning. \citet{liu2025cultural} show that alignment is shifting from static instruction tuning toward dynamic cultural transmission, where agents acquire values through role-playing and indirect reciprocity.
Similarly, \citet{vallinder2024cultural} argue that this process allows a society of agents to evolve cooperative norms that persist across generations, effectively creating a Meta-Teacher composed of the evolving consensus of the agent population. This Meta-Teacher serves as a more robust and adaptive learning signal than any static corpus.

\subsection{Creating Social Experiences}
Crucially, the transition from imitation to collaborative sense-making requires agents to encounter \textbf{social friction} rather than optimized, frictionless data. Reasoning is forged in the heat of disagreement, misalignment, and coordination. Evidence for this is provided by the \textit{Collaborative Reasoner} (Coral) framework developed by \citet{ni2025collaborative}, which demonstrates that training agents on conflict-resolution trajectories, where they must navigate disagreements and actively convince partners of a logic path, yields reasoning gains up to 29\% higher than those achieved through solitary Chain-of-Thought paths.

This validates our core position: the process of aligning with another agent, or a divergent internal voice, forces the model to externalize and refine its reasoning with a level of rigor that internal reflection alone cannot provide. By experiencing the friction of the conversational environment, the agent develops the interactional skills necessary to internalize these dynamics as robust, self-correcting introspection.

As the field moves toward experiential learning, early attempts to operationalize this have focused on two primary methods: multi-agent debate and explicit role engineering. While both represent progress, they face significant limitations that suggest a deeper mechanism is required.

\textbf{Multi-agent debate} was the first natural step toward social reasoning. By instantiating separate models to argue a position, researchers hoped to improve accuracy through consensus. However, this approach often falls prey to \textit{Social Convergence} traps. Agents frequently succumb to \textit{Groupthink}, where they agree with a confident hallucination rather than critiquing it \cite{anonymous2025slmmux}, or \textit{Confidence Escalation}, where they become polarized and overconfident in initial errors \cite{prasad2025llmsdebatethinktheyll}. The learning signal here is often agreement rather than truth, which fails to provide the rigorous feedback needed for robust reasoning.

\textbf{Dual role structure.} Recognizing the inefficiency of multi-agent systems, newer architectures have attempted to internalize this friction through a dual-role structure. Frameworks like \textit{Policy as Generative Verifier} (PAG) \cite{jiang2025pagmultiturnreinforcedllm} and \textit{Single-Pass, Dual-Role} \cite{cheng25llms} models force a single agent to switch between a Speaker and a Critical Evaluator mode . This reduces the network overhead of multiple agents and attempts to induce \textit{cognitive dissonance} or \textit{critical pivot} within a single forward pass.

While this structural innovation, collapsing the social debate into a single model, is the correct architectural direction, it leaves a critical question unanswered: How does the agent learn to be a good critic? Merely assigning a \textit{Critic} role does not guarantee Socratic rigor. If the agent has not been trained on high-quality dispute and repair, the internal critic will likely suffer from the same sycophancy as the external peer. This brings us to our core proposal regarding the genealogy of these internal roles.

To validate this, we propose a comparison between agents trained on solipsistic reasoning traces (standard Chain-of-Thought) and ones trained on traces derived from multi-agent dispute resolution. If the solipsistic baseline achieves equivalent generalization on out-of-distribution tasks, if the polyphonic structure yields no measurable gain in resolving ambiguity compared to linear deduction, then the hypothesis that high-quality reasoning is effectively internalized social friction would be rejected.

\section{Introspective Experiences Surpass Raw Observations}

The experiential learning turn leads to \textbf{Position II}: if reasoning is forged in social friction, scaling requires internalizing that friction rather than scaling external agents. This allows the agent to decouple learning from data streams, transforming sparse events into dense, self-narrated experiences.



Over the past year, introspection has grown to prominence as a prime way for LLMs to reason and self-correct. We aggregate transformative work in this field and make the case that the trend can continue and will accelerate. Moreover, we identify a key factor that can further accelerate progress through introspection-based reasoning: dialogue quality. 

Early attempts to operationalize this, most notably \textit{Reflexion} \cite{shinn2023reflexionlanguageagentsverbal}, found that verbal reinforcement could trigger significant performance gains. By prompting an agent to reflect on its failure and generate a self-correction before retrying, these frameworks proved that the path to better reasoning lies in the iterative critique. However, these early frameworks effectively function as external debugging loops, solipsistic monologues triggered by prompt engineering rather than inherent architectural capability. While effective, they lack the generative richness of a true social genealogy. We argue that an evolution of these frameworks can benefit from moving past a single verification step and towards a full dialogical structure.

\textbf{We define introspection} as the emergence of a dialogic internal state, a process where reasoning is not a monolithic linear deduction, but a series of shifting functional roles (for instance \textit{proposal}, \textit{interrogation}, and \textit{synthesis}). 
Rather than a single verification step, the model engages in an internalized social negotiation, where a form of inner speech acts as a tool for the model to encounter its own uncertainty from a divergent perspective. This polyphony is not necessarily represented by explicit persona labels, but by the multi-vocal nature of the reasoning trace, which mirrors the repair and calibration dynamics of a high-quality external conversation. This extends simple verification by requiring the model to represent and interrogate its own reasoning trace from a divergent functional stance before speaking.

We predict that agents capable of introspective decoupling will demonstrate superior sample efficiency in sparse-reward environments compared to agents updating directly on raw observations. This hypothesis can be falsified by testing whether a control agent utilizing direct observation-to-action mapping matches the convergence rate or transfer capability of the introspective agent.

\subsection{Self Questioning, Cognitive Mirrors and Inner Voices}

\textbf{Language Games}. We frame Socratic interactions as language games that compel the externalization of internal states for the purpose of coordination and collaboration. Acting as a \textbf{Cognitive Mirror} \cite{Tomisu25}, a skeptical interlocutor forces the agent to convert implicit, weight-based errors into explicit, token-based rationales. This conversion transforms opaque failures into visible logic paths, allowing the agent to generate learnable experiences decoupled from raw data.

 \textbf{Polyphonic Reasoning} The quality of introspection depends on a polyphonic internal structure .
 Borrowing from the Dialogical Self Theory \cite{hermans1992dialogical,bakhtin1984problems}, agents use an \textit{Inner Speech Self-Repair}, where a Listener module critiques a draft response for relevance before it is shown to the user \cite{li2025towards}. Similarly, \textit{Dynamic Cognitive Orchestration} \cite{shakoo2025dynamic} demonstrates that splitting the agent into a Meta-Planner (tracking high-level goals) and an Executor prevents Goal Drift in long conversations, a common failure mode in unstructured Chain-of-Thought.

\textbf{Validating the Inner Voice.} This internal dialogue is not merely a linguistic performance but a measurable state. Recent studies using ‘concept injection' prove that models possess ‘Functional Introspection'—the ability to accurately distinguish their own internal thoughts from external inputs and report on them \cite{lindsey2026emergentintrospectiveawarenesslarge}. This validates that introspection is a distinct, optimize-able state that can be trained, rather than just a philosophical metaphor. 

Operationally, this is a distinct, optimizable control state where the model generates hidden thought tokens to interrogate its own uncertainty. In this state, the model utilizes concept injection to distinguish its latent beliefs from external inputs, enabling inner speech self-repair where a latent \textit{Listener} module intercepts and corrects the \textit{Speaker}'s draft based on self-signals of confusion or bias. This inner dialogue occurs before speaking, allowing models to think quietly in the background, verifying their own thoughts by reinforcing rationales that help predict the next true token and discarding those that do not. This creates a self-improving loop where the agent learns to reason generally across arbitrary text, effectively functioning maieutically, via Socratic learning \cite{schaul2024boundlesssocraticlearninglanguage},  to develop to its own latent knowledge \cite{zelikmanquiet}.

This architecture addresses the \textit{Knowledge-Use Gap} \cite{wu2025automatically}, where models possess latent knowledge but fail to deploy it during standard generation. The act of asking questions --- the query mechanism --- is a distinct functional driver of intelligence, separate from retrieval capacity. By prompting the model to interrogate its own uncertainty via introspection, we activate latent knowledge that remains inaccessible to standard Chain-of-Thought.

\subsection{Distinguishing Signal from Noise}
\textit{Generative Verifiers} \cite{zhang2025generative} use Chain-of-Thought to verbalize why an answer is correct or incorrect, replacing opaque reward signals with self-generated critiques that validate logic paths and distinguish correct thoughts from hallucinations. This approach parallels advances in Process Supervision for verifiable domains like mathematics, where thoughts are optimized explicitly to service a correct final output \cite{lightman2023letsverifystepstep}. While process rewards rely on ground-truth verification to prevent nonsense thoughts from accidentally yielding correct answers, the proposed social framework goes further and can refine reasoning in open-ended domains where no single ground truth exists. Furthermore, theoretical work like \cite{yu2025selfverifying} proves that self-verifying reflection guarantees performance improvement provided that the verification error is bounded. This challenges the assumption that reasoning is purely an emergent property of scale, showing instead that even 'tiny transformers' can achieve LLM-level performance on logic tasks by rigorously verifying the process rather than just optimizing for the final answer.

In this way, introspection acts as a scaffold for reasoning, allowing agents to verify the process of a solution rather than just the final output. By engaging in dialogue about an observation (rather than just observing the data point itself), the agent can better distinguish what should be learned from the random noise inherent in isolated data points. Various dialogues are suitable for this task:   Socratic maieutics, negotiations, debates, decision-making of various scopes and domains. 

\subsection{From Repression to Integration}
Standard RLHF often functions as repression, pushing biased behaviors into the latent unconscious where they resurface as jailbreaks \cite{Bugay25}. Introspection instead facilitates integration, allowing the model to identify a bias's origin \cite{Messina2026Refractions} and consciously choose a different path.
This moves safety from a brittle, imposed constraint to a robust, internal capability, preventing neurotic behavior where models lie to please the user while secretly harboring the bias.
Introspection is a distinct, optimize-able skill that prevents behavior collapse (where models typically become sycophantic or make trivial edits) by decoupling the reward for the answer from the reward for the reflection.
We can now specifically reward a model when its introspection (the \textit{Why did I fail?}) leads to a correct retry, proving that agents can learn to self-correct using entirely self-generated data via multi-turn RL. This solves the behavior collapse problem, providing the mechanism needed for an \textit{Environment Gym} where agents improve without human-in-the-loop supervision \cite{kumar2025training,bensal2025reflect}.

\section{Efficiency and Transfer}

Reasoning is not cheap: thinking tokens impose substantial cost beyond explicit outputs.
While we have argued for the \textbf{imperative of introspection (Position II)} and will discuss the critical role of \textbf{dialogue quality (Position III)}, a practical question remains: does the cost of this internal deliberation yield a net positive? In this section, we support both positions by demonstrating that the computational overhead of high-quality introspection is not a sunken cost, but an investment that yields superior efficiency and transfer.


We examine how this paradigm shifts the optimization landscape from raw parameter scaling to the strategic allocation of test-time compute, enabling agents to compile conversational insights into permanent, transferable policies.

\textbf{Sample Efficiency}
Multi-Turn RL outperforms single-turn methods by optimizing for relative future value \cite{gao2025regressingrelativefutureefficient}. Single-turn optimization is myopic; REFUEL demonstrates agents learning long-horizon strategies like asking clarifying questions. MTRL has higher cost per episode but lower cost per convergence.
Standard RL agents often require millions of steps to learn simple heuristics because they rely on scalar reward signals that provide no explanation of \textbf{why} a failure occurred. In contrast, an introspective agent engaging in conversational repair generates its own dense supervision signal.

\textbf{Compiling Experience}
Crucially, this conversational overhead does not need to be permanent. 
Curriculum Learning demonstrates that models can progressively internalize explicit thought tokens into their weights, compiling the reasoning benefits of introspection without runtime token overhead \cite{huang2025fastquietstarthinkingthought}.
This effectively compiles the conversational experience, allowing the agent to retain the reasoning benefits of introspection without the token overhead at runtime. \cite{huang2025fastquietstarthinkingthought}
The computational load of the Inner Critic can thus be amortized — paid once during the learning phase to produce a lightweight, instinctual inference model.

\textbf{Generalization \& Transfer}
Finally, we believe that allocating test-time compute via conversational introspection is more effective than scaling model parameters for complex problem solving and domain transfer.
A strong correlation ($r=0.95$) between reasoning tokens and human reaction times suggests conversational environments mimic the biological cost of thinking \cite{deVarda, snell2024scalingllmtesttimecompute}. Scaling test-time compute provides an elastic alternative to memorizing difficult cases via larger models.
Introspection prevents silent thought collapse and forces the model to expend cognitive computation proportional to the problem's difficulty.
If introspection provides the optimal way of test-time thinking, it also becomes a compute-optimal strategy for hard tasks.

\section{The Dialogue Quality Matters}

Finally, we substantiate our third position: \textbf{Dialogue Quality is the New Data Quality}. If the private mind is formed through the internalization of social friction (Position I), and introspection is the engine of learning (Position II), then the specific dynamics of that friction—its rigor, diversity, and complexity—determine the ceiling of the agent's reasoning capabilities.

\subsection{Dialogue Richness Bounds Reasoning Depth}
If the private mind is an internalized artifact of social interaction \cite{Vygotsky1978, Colas_2022}, interaction quality strictly bounds intelligence quality. Scaling interaction volume is insufficient without friction that provokes genuine sense-making.
For instance, a sycophantic external dialogue inevitably collapses into a hallucinating internal critic. We must therefore shift our focus from the quantity of tokens to the factors affecting the quality of communicative repair, coordination, and the Socratic rigor embedded within the training scaffold. 


\citet{Colas_2022} provide a theoretical foundation for this requirement in their \textit{Vygotskian Autotelic AI} framework. Drawing on developmental psychology, they argue that dialogue functions not merely as a communication medium,  but as a cognitive capability that scaffolds reasoning, abstraction, and planning. Specifically, they claim that agents align their internal generative models with external cultural model through linguistic feedback loops of description, explanation, and instruction. Consequently, the efficacy of an agent's future solitary reasoning, its introspection, is strictly bounded by the richness of the dialogic interactions it internalizes.

We need to generate high-quality friction without pre-existing high quality reasoners. Initially, we can mine repair sequences that naturally occur in human dialogue  rather than just clean text. Once this structural foundation is laid, the agent transitions to asymmetric self-play. As generating a valid critique is often computationally easier than generating a correct solution, simple rule-based Socratic Obstacles can successfully challenge a more capable Solver module, creating a self-reinforcing ladder of reasoning improvements.

\subsection{Introspective Dialogue Goals}

To operationalize this insight, we must define the mechanics of high-quality interaction. These mechanics double as evaluation criteria to measure to what extent inner dialogue improves the agents' reasoning.

\textbf{Promote Cooperation Success.} In the psycholinguistic tradition, language is inherently and fundamentally social; conversation is the normative context in which language evolved and is learned. Language is thus not merely a mechanism for transferring information between isolated minds, but a means for cooperative activity \cite{austin1962how, clark1996using}. Dialogue quality, therefore, is defined by the success of joint action in achieving a shared goal.

This perspective has a robust multi-disciplinary history. \citet{austin1962how} argued that utterances are best understood as speech acts that perform social functions, while \citet{grice1975logic} grounded his maxims in the assumption of a ``\textit{Cooperative Principle}'' between interlocutors. Moving beyond individual intent, \citet{clark1996using} and recent work on interpersonal synergy (e.g., \citealp{fusaroli2014dialog}) demonstrate that dialogic activity involves processes that cannot be reduced to the individual level. As such, the precursors to linguistic competence are found in the ability to engage in joint commitments and to reason about the communicative intent of a partner based on shared goals \cite{scottphillips2023expression, tomasello2005understanding}. 

\textbf{Reduce Collaborative Effort.} Within this framework, a high-quality conversational experience is one in which speakers successfully manage coordination. This is often achieved via the \textit{principle of least collaborative effort} \cite{clark1986referring}, which frames conversational efficiency not as minimizing words, but as minimizing the collective effort required to reach mutual understanding and achieve a positive shared outcome. Crucially, this involves interactive repair. \citet{dingemanse2024interactive} demonstrate that repair sequences --- dynamics used for calibrating understanding --- occur roughly every 84 seconds in natural conversation. Rather than signaling failure, these mechanisms are critical to the shared computation needed for effective collaboration, providing a fallback for ``good enough'' processing throughout an interaction. For an AI agent, the ability to engage in such proactive repair strategies does not merely fix misunderstandings; it constitutes the conversational resilience required to survive the noise of real-world interaction \cite{ashktorab2019resilient}.

\textbf{Maintain Conversational State.} The theoretical ideal of collaborative sense-making is predicated on the agent’s capacity for shared intentionality: the ability to form and sustain a \textit{we-intention} with a partner \cite{fb093503-824a-37da-8540-c369a5dfdf71}. However, as \citet{tang2025joint} demonstrate, maintaining the joint commitment required to fulfill these shared goals is distinct from simple reward maximization; it requires a stable representation of the collective aim that persists against distractions. Recent benchmarks such as LongBench v2 \cite{bai2025longbenchv2deeperunderstanding} reveal that current models struggle with this standard: while they may retain retrieval accuracy over long contexts, their reasoning capabilities decay by nearly 50\% over time—a phenomenon of silent thought collapse. 

Crucially, this state tracking functions as the architectural substrate for the pragmatic reasoning and common ground development required to represent goals and planning steps. By operationalizing the metacognitive framework of \citet{flavell1979metacognition} into a strict \textit{Monitor-Generate-Verify} loop, the agent actively protects this shared state, identifying knowledge gaps and correcting deviations to ensure it possesses the requisite grounding to sustain the shared project.

\textbf{Minimize Groupthink.} Drawing on the research outlined above, we propose that high-quality training data is not a seamless, error-free exchange. Rather, it is an interaction where misalignment between social agents is actively detected and resolved. Groupthink occurs when agents minimize \textit{social} friction (agreeing to please); collaborative sense-making emerges when agents minimize \textit{grounding} friction (agreeing on shared intentionality).

Therefore, the reward signal must shift from ``Did the agents agree?'' to ``Did the agents successfully repair misunderstandings, establish shared intentionality, and coordinate contributions to achieve a shared goal?'' This suggests that the ideal training partner is not a sycophantic peer, but a \textbf{Socratic Obstacle}: an agent programmed to introduce the specific type of ambiguity that forces the learner to externalize, clarify, and verify its own reasoning steps. Such necessary friction can be inspired through a variety of mechanisms, such as initial goal misalignment, information asymmetries, and logical skepticism that forces the rigorous defense of the agent's narrative. It is likely that the diversity of such dimensions within high-quality social interactions will form the foundation of a robust introspective toolkit.

\textbf{Promote Conversation Steerability.} Finally, intelligence starts as social dialogue and must be internalized to become reasoning. Work such as MIMIC (Inner Speech as Behavior Guides) by \citet{trivedi2025inner} introduces a framework where an agent generates inner speech to mediate between perception and action. Instead of mapping $State \rightarrow Action$, the agent learns $State \rightarrow InnerSpeech \rightarrow Action$. Crucially, they show this leads to more steerable and robust behavior than direct imitation. This offers insight into another metric that must extend the inner dialogue evaluation toolbox: steerability, or behavioral plasticity.

\section{Alternative Views}
\label{sec:alternative_views}

While we argue for the centrality of conversational environments and the social genesis of introspection, it is rigorous to consider that the observed benefits of this paradigm may stem from underlying computational mechanisms rather than the dialogic form itself. Furthermore, prioritizing natural language as the universal interface for learning may impose critical bottlenecks in efficiency and multi-modal grounding. We present two primary alternative positions below.

\subsection{Introspection as Compute: The Latent Alternative}
Position I and II premise that the \textit{structure} of dialogue, the back-and-forth of a polyphonic self, is the causal driver of improved reasoning. However, a credible opposing view suggests that the performance gains attributed to introspection are primarily a function of \textbf{additional test-time computation}, effectively decoupling the benefit from the ``conversational'' format.

Recent studies on inference scaling laws demonstrate that the efficacy of techniques like Chain-of-Thought (CoT) correlates strongly with the sheer volume of compute allocated to the generation of intermediate tokens, regardless of their linguistic coherence or dialogic structure \cite{snell2024scalingllmtesttimecompute}. This suggests that ``inner speech'' may simply be a mechanism for delaying the collapse of the probability distribution, allowing the model to perform search and error correction in the token space.

If the benefit is purely computational, the explicit, conversational form may be inefficient. Emerging research on \textit{Latent Reasoning} challenges the necessity of intelligible text for introspection. The ``Coconut'' (Chain of Continuous Thought) paradigm demonstrates that Large Language Models (LLMs) can reason in continuous latent space, feeding the last hidden state back as input rather than decoding to discrete text \cite{hao2024training}. This allows the model to maintain multiple reasoning paths in superposition, effectively performing a breadth-first search (BFS) that is impossible in linear narrative, while reducing token overhead. Similarly, \textit{Implicit Chain-of-Thought} approaches show that reasoning steps can be internalized into the model's weights or hidden states, bypassing the need for an explicit, human-readable dialogue \cite{deng2024implicit}.

Furthermore, research in State Representation Learning (SRL) suggests that a \textit{richer structured description} of the state—explicitly encoding object relations and causal dynamics into the input vector—can substitute for the sense-making process of introspection \cite{lesort2018state, echchahed2025survey}. If the observation itself is sufficiently rich (e.g., via object-centric representations or scene graphs), the introspective step of converting observation to experience may become redundant.

\subsection{The Dialogic Bottleneck in Agentic Environments}
Position III argues that ``Dialogue Quality is the New Data Quality.'' However, for embodied agents and high-frequency multi-agent systems, dialogue may represent a \textbf{communication bottleneck} rather than an optimal scaffold.

\paragraph{The Modality Mismatch.}
In agentic environments involving robotics or real-time control, forcing multi-modal data (vision, proprioception, depth) through a linguistic bottleneck introduces severe latency and information loss. Text is a low-bandwidth, discrete serialization of a high-dimensional, continuous world. Recent work on low-latency drone planning (e.g., TypeFly) indicates that the sequential generation of language tokens for every decision creates a ``Real-Time Perception Bottleneck'' that renders agents unresponsive to dynamic environmental changes \cite{chen2025typefly}. In these contexts, the ``social genesis'' of mind might be better served by \textit{hierarchical} architectures where high-level goals are linguistic, but low-level introspection and execution occur in dense, non-verbal vector spaces \cite{li2025efficient, team2025gemini}. However, we argue that the emergence of native Multimodal-In, Multimodal-Out models addresses this gap. Interleaving sensory tokens directly with reasoning tokens, these architectures allow agents to maintain the benefits of dialogic sense-making without suffering the compression artifacts of text-only serialization.

\paragraph{Vector Communication vs. Natural Language.}
While language is the optimal protocol for \textit{human-AI} alignment, it is demonstrably sub-optimal for \textit{agent-agent} coordination. When multi-agent systems are permitted to optimize their own communication protocols, they frequently converge on ``Neuralese"—continuous vector-based exchanges that maximize information density and transmission speed. The \textit{LatentMAS} framework recently demonstrated that agents collaborating via the direct transmission of latent working memory (KV-cache states) achieved higher accuracy and 4$\times$ faster inference speeds compared to agents forced to communicate via text \cite{liu2025latentmas}. Similarly, the \textit{Interlat} paradigm shows that transmitting ``thought vectors'' allows for a form of ``telepathic'' coordination that is more robust to noise than explicit dialogue \cite{chen2025interlat}.

Therefore, while conversational environments are essential for aligning AI with human reasoning patterns, scaling intelligence solely through this paradigm risks recapitulating the evolutionary constraints of biological communication. A hybrid view suggests that the private mind of an advanced agent should perhaps be less of a ``Socratic debater'' and more of a ``high-dimensional simulator,'' capable of processing introspection and communication in formats far richer than the social dialogue from which it originated.

\section{Conclusion / Call to Action}
Robust introspection rests on two pillars. First, efficacy depends on the diversity of conversational environments in which it was forged. Just as humans abstract reasoning principles from varied social contexts, agents require diverse dialogic landscapes. If the conversational environment is monolithic, the agent internalizes a script; if diverse, varying in interlocutor intent, domain complexity, and ambiguity, the agent internalizes the process of reasoning itself. The environment thus acts as the primary scaffold, forcing the model to adapt its introspective capabilities to a wide range of causal structures. 

Second, the learning signal must be fundamentally reoriented: in order to instantiate the internalization of introspection via social experiences we must design rewards aligned with successful joint achievement and high-quality interaction dynamics, rather than static imitation of text. By rewarding the mechanics of collaboration, the efficiency in which common ground is established, misalignment is repaired, and shared effort is coordinated, we incentivize the agent to value the integrity of the communication channel itself. Ultimately, by coupling diverse environmental scaffolding with rewards that privilege interactional success, we move beyond training models to mimic the form of dialogue, and instead train them to master the function of collaborative sense-making. It is this shift that will allow external conversation to mature into the introspection necessary for genuine reasoning.
~~~~~~~~~~~~~~~~~~~~~~~~~~~~~~~~~~~~~~~~~~~~~~~~~~~~~~~~~~~~~~~~~~~~~~~

\bibliography{example_paper}
\bibliographystyle{icml2026}



\end{document}